\documentclass{bmvc2k}
\usepackage{graphicx}
\usepackage{amsmath}
\usepackage{amssymb}
\usepackage{booktabs}
\usepackage{mathtools}
\usepackage{multirow}
\usepackage{hhline}
\usepackage{makecell}
\usepackage{tabularx}
\usepackage{wrapfig,booktabs}
\usepackage{footnote}

\DeclareMathOperator*{\argmax}{arg\,max} 

\title{Temporal-aware Hierarchical Mask Classification for Video Semantic Segmentation}

\addauthor{Zhaochong An}{zhaoan@ethz.ch}{1}
\addauthor{Guolei Sun}{guolei.sun@vision.ee.ethz.ch}{1*}
\addauthor{Zongwei Wu}{zongwei.wu@uni-wuerzburg.de}{2}
\addauthor{Hao Tang}{hao.tang@vision.ee.ethz.ch}{1}
\addauthor{Luc Van Gool}{vangool@vision.ee.ethz.ch}{1,3}

\addinstitution{
 Computer Vision Lab,\\
 ETH Zurich,\\
 Zurich, Switzerland
}
\addinstitution{
 Computer Vision Lab, CAIDAS \& IFI,\\
 University of Wurzburg,\\
 Wurzburg, Germany
}
\addinstitution{
 VISICS,\\
 KU Leuven,\\
 Leuven, Belgium\\
}

\runninghead{An, Sun, Wu, Tang, Van Gool}{THE-Mask}

\def\eg{\emph{e.g}\bmvaOneDot}

\def\ie{\emph{i.e}\bmvaOneDot}

\def\ourmodel{THE-Mask}

\begin{document}

\maketitle
\begin{abstract}
Modern approaches have proved the huge potential of addressing semantic segmentation as a mask classification task which is widely used in instance-level segmentation.
This paradigm trains models by assigning part of object queries to ground truths via conventional one-to-one matching. However, we observe that the popular video semantic segmentation (VSS) dataset has limited categories per video, meaning less than 10\% of queries could be matched to receive meaningful gradient updates during VSS training. 
This inefficiency limits the full expressive potential of all queries.
Thus, we present a novel solution \textbf{\ourmodel}~for VSS, which introduces \textbf{t}emporal-aware \textbf{h}i\textbf{e}rarchical object queries for the first time. Specifically, we propose to use a simple \textbf{two-round matching mechanism} to involve more queries matched with minimal cost during training while without any extra cost during inference. To support our more-to-one assignment, in terms of the matching results, we further design a \textbf{hierarchical loss} to train queries with their corresponding hierarchy of primary or secondary.
Moreover, to effectively capture temporal information across frames, we propose a \textbf{temporal aggregation decoder} that fits seamlessly into the mask-classification paradigm for VSS.
Utilizing temporal-sensitive multi-level queries, our method achieves state-of-the-art performance on the latest challenging VSS benchmark VSPW without bells and whistles. The code is available at  \href {https://github.com/ZhaochongAn/THE-Mask}{github.com/ZhaochongAn/THE-Mask}.
\end{abstract}

\section{Introduction}
Video semantic segmentation (VSS) is to assign per-pixel semantic categories to each frame of a video. As a fundamental task of scene understanding, VSS has significant implications for wide applications such as image editing~\cite{hong2018learning}, autonomous driving~\cite{geiger2012we}, and medical diagnosing~\cite{zhou2018unet++}. Besides, the recent release of the large-scale dataset VSPW~\cite{miao2021vspw}, with its higher annotated frame rate, has further spurred advancements in VSS.

VSS can be seen as extending image semantic segmentation to the video domain. Unlike images, videos contain crucial temporal semantic information, which helps segment objects despite their motion blur and occlusions across frames. 
Therefore, simply applying image-level segmentation models~\cite{kundu2016feature,huang2018efficient} to videos yields suboptimal performance. Several  methods~\cite{lei2020blind,zhang2022auxadapt,park2022real} adapt frame-wise predictions for high temporal consistency, while others use per-clip approaches~\cite{miao2021vspw,sun2022coarse,guan2021domain,xing2022domain} trained with video clips to aggregate temporal features. 
Despite these efforts, VSS has been mainly treated as a per-pixel classification task. 
However, recent works~\cite{cheng2021per,cheng2022masked,li2022video} show that mask-classification architecture, commonly used in instance segmentation, can also achieve satisfactory results for semantic segmentation. For example, Mask2Former~\cite{cheng2022masked} follows DETR~\cite{DBLP:conf/eccv/CarionMSUKZ20} to learn object-centric representations as queries that are transformed into final image segments. These representations~\cite{locatello2020object,DBLP:conf/eccv/CarionMSUKZ20} have shown their effectiveness in capturing object features for image input. The insight inspires us to further explore the potential of mask-based approaches for the VSS task.

\begin{figure}[!t]
  \begin{minipage}[c]{0.49\textwidth}
    \includegraphics[width=\textwidth]{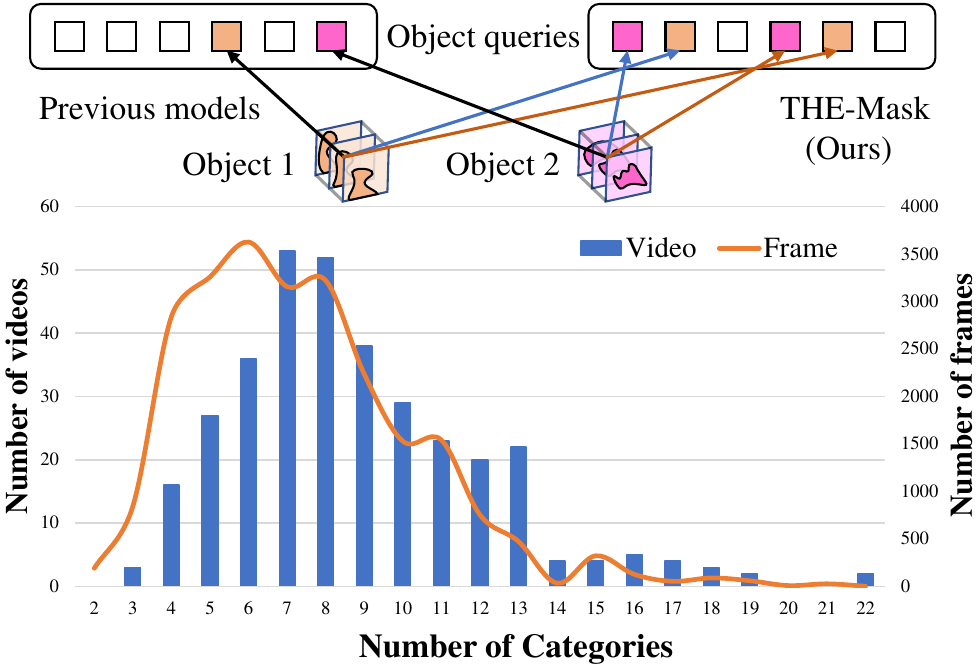}
  \end{minipage}\hfill
  \begin{minipage}[c]{0.48\textwidth}
    \caption{In VSPW, each frame/video has about 8 categories on average. This means only 8 out of 100 queries in previous mask-classification-based models (\eg, Mask2Former~\cite{cheng2022masked}) can be matched to train effectively in one iteration, which limits fully utilizing the expression ability of all queries. By contrast, our \ourmodel~can involve 2$\times$ matched queries with rich semantic hierarchy for better performance.} \label{fig:data}
  \end{minipage}
  \vspace{-17pt}
\end{figure}

In this work, we rethink the application of the mask classification paradigm in VSS. This paradigm predicts for each learned object query a pair of a binary mask and a class distribution. The final segmentations are aggregated from all predicted pairs. By default, the number of object queries is usually set to be a large number (\eg, 100 in~\cite{cheng2022masked}). During training, most existing methods~\cite{DBLP:conf/eccv/CarionMSUKZ20,cheng2021per,cheng2022masked,li2022video} adopt a one-to-one bipartite matching, \ie, Hungarian matching~\cite{kuhn1955hungarian}, to assign the best-fit predicted pair to each ground truth pair. Regarding the loss, the matched predicted pairs are optimized with the ground truths, while the unmatched pairs are forced to predict the artificial “no-object” category. In such a manner, only matched queries can be trained effectively while unmatched queries receive meaningless updates. However, as shown in Fig.~\ref{fig:data}, in the benchmark VSPW dataset, only about 8\% queries can be matched on average to receive informative gradient updates for each training sample. This hinders fully utilizing the representation ability of all queries and harms performance. Given this observation, one question emerges: \textit{is there a simple method to involve more queries during training without losing their own object representation abilities?} 
Besides, to leverage the temporal consistency in the frame domain, the second question naturally raises: \textit{How to effectively model the temporal interactions in the mask-classification-based paradigm?}

To address the above questions, we propose a novel framework, termed \textbf{\ourmodel}, for VSS. For the first question, we introduce a \textbf{two-round matching mechanism} to bind more queries with ground truths, yielding a more-to-one matching manner for VSS. To support this paradigm, we propose a \textbf{hierarchical loss} which enables each query to learn its specific non-overlapping object representation. 
This increases the expression variety of matched object queries and results in hierarchical queries with rich semantics at different levels. These complementary queries contribute to the final segments from both primary and secondary views, leading to improved performance. 
For the second question, we propose a lightweight \textbf{Temporal Aggregation Decoder (TAD)} that utilizes two sets of queries, \ie, video-level and image-level, to explicitly model image-video relationships and learn temporal information, optimizing the interactions between queries and multiple frames. TAD effectively avoids hard long attention sequences while adding only 3.2M parameters over the baseline counterpart.

We evaluate \ourmodel~against the state-of-the-art methods on the most challenging VSS dataset VSPW~\cite{miao2021vspw}. Our \ourmodel~trained on 4-frame clips ($t=4$) achieves the state-of-the-art result of 49.1\% mIoU (MiT-B2~\cite{xie2021segformer} as Backbone), surpassing the prior method MRCFA~\cite{sun2022mining} by 3.8\% mIoU.
Compared to our baseline~\cite{cheng2022masked}, our method achieves a 0.9\% absolute gain in mIoU ($t=1$) without any extra inference cost and further achieves better performance with longer clips. We also perform a detailed ablation study to validate the effectiveness of our approach. Our experiments demonstrate the efficacy of \ourmodel~and highlight the potential of mask-classification models for the VSS task. We believe that our model is an effective baseline for future VSS research. In summary, the contributions of \ourmodel~are as follows:
\vspace{-5pt}
\begin{itemize}
\setlength\itemsep{1pt}
\item  Our \ourmodel~is the first of its kind to introduce hierarchy into object queries in the mask-classification-based paradigm. The parameter-free hierarchical design enriches the expression ability of queries and improves the performance without any extra inference cost. 
\item To leverage the temporal clues in the video setting, we propose a temporal aggregation decoder to effectively model cross-frame interactions while fitting seamlessly into the mask-classification paradigm. \ourmodel~achieves the new state-of-the-art performance on the VSPW benchmark.
\end{itemize}

\vspace{-10pt}
\section{Related Work}
\textbf{Image semantic segmentation (ISS)}
is to assign a semantic label to each pixel of the input image. Naturally, it can be formulated as a per-pixel classification task. From the early FCNs~\cite{long2015fully}, most works follow the per-pixel setting and differ to exploit semantic information in proposing new structures to enlarge the receptive field~\cite{DBLP:journals/corr/ChenPKMY14, yu2015multi, noh2015learning, dai2017deformable};  designing multi-scale feature ensemble methods~\cite{liu2015parsenet,hariharan2015hypercolumns,ronneberger2015u, chen2016attention, xia2016zoom, zhao2017pyramid, chen2017deeplab, chen2017rethinking, lin2017refinenet, yang2018denseaspp, chen2018encoder,he2019dynamic}; aggregating stored dataset-level representations~\cite{jin2021mining}; using non-local context aggregation schemes~\cite{zhao2018psanet,fu2019dual,sun2020mining,yuan2021ocnet,jin2021isnet,hoyer2022daformer}; or utilizing long-range modeling capacity of transformers~\cite{strudel2021segmenter,xie2021segformer,zheng2021rethinking}. More recently, motivated by DETR~\cite{DBLP:conf/eccv/CarionMSUKZ20}, MaskFormer~\cite{cheng2021per} and Mask2Former~\cite{cheng2022masked} address ISS using query-based transformer architectures with a mask-classification-based paradigm, which has been widely used in instance-level segmentation~\cite{he2017mask, cai2018cascade,chen2019hybrid,tian2020conditional,wang2020solov2,cheng2020panoptic,wang2021max,DBLP:journals/corr/abs-2112-08275}. 
The success of the mask-classification-based perspective for ISS inspires us to explore the new paradigm in the video domain.

\noindent\textbf{Video semantic segmentation (VSS)}
as an extension of ISS aims to predict pixel-level semantics in consecutive video frames. 
Some works treat VSS in a per-frame fashion and refine the predictions for temporal consistency~\cite{kundu2016feature,hur2016joint,huang2018efficient,lei2020blind,zhang2022auxadapt,park2022real}.
Other works explore different mechanisms in training to fuse semantic information across time by using patch matching or optical flow for label propagation~\cite{badrinarayanan2010label,mustikovela2016can,budvytis2017large}; utilizing the predictive learning features carrying the temporal context~\cite{luc2017predicting,jin2017video}; employing recurrent units to propagate features from past frames to current frame~\cite{fayyaz2016stfcn,nilsson2018semantic}; exploiting aligned previous segmentation maps as supervised signals~\cite{zhu2019improving,liu2020efficient,guan2021domain,xing2022domain}; aggregating the dataset-level representations of previous frames~\cite{jin2022mcibi++}; or warping representations of adjacent frames~\cite{gadde2017semantic,liu2017surveillance,li2018low,ding2020every,miao2021vspw,sun2022coarse,sun2022mining} by using various temporal-adapted wrapping modules~\cite{horn1981determining,yuan2020object,zhao2017pyramid}. 
Beyond exploring more temporal-accurate models, another line of research focuses on improving the efficiency~\cite{zhu2017deep,mahasseni2017budget,xu2018dynamic,jain2019accel,liu2020efficient,hu2020temporally}. 
Recently, a few works have applied query-styled approaches in the video domain as well.~\cite{kim2022tubeformer} built upon mask transformers~\cite{wang2021max} introduces a latent memory to facilitate attention learning over multiple frames. Video k-net~\cite{li2022video} adapts K-net~\cite{zhang2021k} into the video by performing kernel interactions along the temporal dimensions.

\begin{figure*}[t]
    \centering
    \includegraphics[width=.98\linewidth]{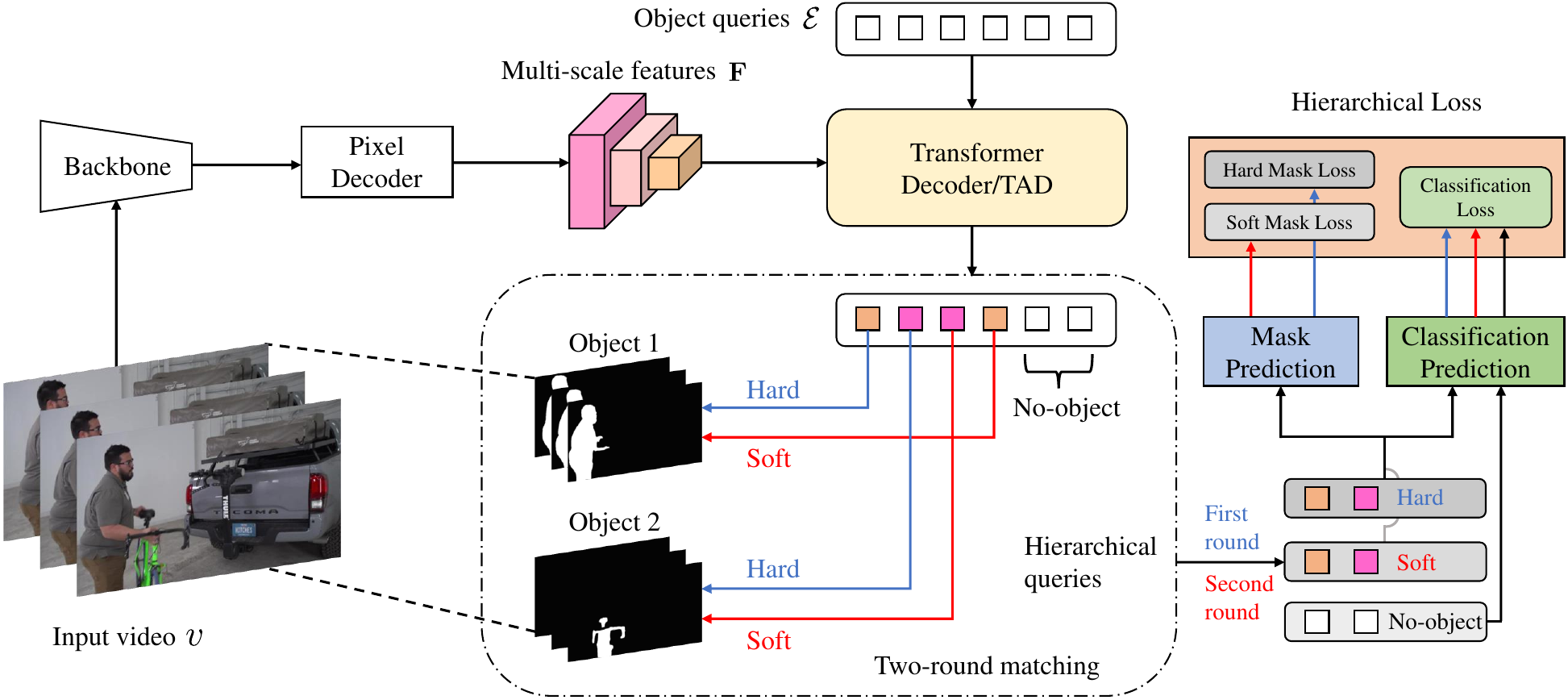}
    \caption{\textbf{\ourmodel} uses the mask-classification-based architecture with a backbone, a pixel decoder, and a transformer decoder. To fully exploit the expression ability of queries, we propose a hierarchical loss, which can effectively train 2$\times$ queries as before in one training iteration with richer semantic features. Beside, to learn the temporal clues, we design the Temporal Aggregation Decoder (TAD) to replace the original transformer decoder.}
    \label{fig:dis}
    \vspace{-10pt}
\end{figure*}

\section{Proposed Method}
\label{sec:method}
\subsection{Preliminaries}
\label{section:pre}
\textbf{Mask classification formulation.} Given an input video $v \in \mathbb{R}^{T \times 3 \times H \times W}$ with $T$ frames of spatial size $H \times W$, the traditional VSS models predict the probability distribution over all categories $\mathbb{C} = \{1, ..., K\}$ for each video pixel: $\{p_i|p_i \in \Delta^K\}_{i=1}^{T \cdot H \cdot W}$. Here, $\Delta^K$ is the $K$-dimensional probability simplex. Different from the above per-pixel classification manner, we use the mask classification paradigm, which learns a set of $N$ queries $\mathcal{E} = \{q_1, ..., q_N| q_i \in \mathbb{R}^C\}$ as object-centric representations to output $N$ classification-mask pairs: 
\begin{equation}
    \{\hat{y}_i\}_{i=1}^N = \{(\hat{p}_i, \hat{m}_i)| \hat{m}_i \in \{0,1\}^{T \times H \times W}, \quad \hat{p}_i \in \Delta^{K+1}\}_{i=1}^N,
\end{equation}
where $\hat{m}_i$ is the predicted mask from object query $q_i$, and $\hat{p}_i$ containing $K$ categories and a "no object" label ($\varnothing$) is the class probability distribution assigned to $\hat{m}_i$. 
Let us denote $\hat{p}_i(c)$ as the probability of assigning class $c \in \mathbb{C}$ to mask $\hat{m}_i$. During inference~\cite{cheng2021per,cheng2022masked}, it will aggregate all the predicted pairs $\{\hat{y}_i\}_{i=1}^N$ to the segmentation output $\hat{y}_{t,h,w}$ for pixel $(h,w)$ at frame $t$ by: $\hat{y}_{t,h,w} = \argmax_{c\in \mathbb{C}} \sum_{i=1}^N \hat{p}_i(c)\cdot \hat{m}_{i,t,h,w}$.

Mask classification models firstly use a \textit{backbone} to extract the features of each input frame individually, then refine the features via a \textit{pixel decoder}, and finally apply a \textit{transformer decoder} to learn the object queries and output $\{\hat{y}_i\}_{i=1}^N$ in parallel. 
Training such models need a one-to-one bipartite matching $\sigma = \{\sigma(i)\}_{i=1}^{N^{gt}}$~\cite{kuhn1955hungarian,stewart2016end,DBLP:conf/eccv/CarionMSUKZ20} to assign the best-fit $N^{gt}$ predictions from $\{\hat{y}_i\}_{i=1}^N$ to the ground truth set $\{y_i = (c_i, m_i)\}_{i=1}^{N^{gt}}$, where $c_i^{gt} \in \mathbb{C}$ is the ground truth class label of mask $m_i \in \{0,1\}^{T \times H \times W}$. Thus, we denote $\hat{y}_{\sigma(i)}$ as the matched pair to $y_i$. The matching score $\mathcal{S}_{match}$ used for evaluating the fitness between $\hat{y}_j$ and $y_i$ is defined as:
\begin{equation}
    \label{eq:smatch}
    \mathcal{S}_{match}(\hat{y}_j, y_i) = \underbrace{\lambda_{ce}\mathcal{L}_{ce}(\hat{m}_j,m_i) + \lambda_{dice}\mathcal{L}_{dice}(\hat{m}_j,m_i)}_\text{$\mathcal{L}_{mask}$} - \hat{p}_j(c_i),
\end{equation}
where $\mathcal{L}_{ce}$ is the binary cross-entropy loss and $\mathcal{L}_{dice}$ is the dice loss~\cite{milletari2016v} with weights $\lambda_{ce}$, and $\lambda_{dice}$. The remaining $N-N^{gt}$ queries are matched to $\varnothing$ class. Then the optimization of model parameters $\theta$ is via minimizing the final loss $\mathcal{L}$ over all queries, consisting of $\mathcal{L}_{mask}^\sigma$ over matched queries in $\sigma$ and a classification term $\mathcal{L}_{cls}$ for unmatched queries to predict $\varnothing$ and for matched queries to ground truth labels:
\begin{equation}
\begin{split}
    &\min_{\theta}\mathcal{L} = \sum_{i=1}^{N^{gt}} \mathcal{L}_{mask}^\sigma(\hat{m}_{\sigma(i)}, m_i) +\underbrace{(-\sum_{i=1}^{N^{gt}}\log\hat{p}_{\sigma(i)}(c_i)-\sum_{i\notin \sigma}\log\hat{p}_i(\varnothing))}_\text{$\mathcal{L}_{cls}$}.
\end{split}
\end{equation}

\subsection{Hierarchical Mask Classification}
\label{section:hq}
To involve more queries into training, we design an effective two-round matching with hierarchical loss functions to leverage richer expression ability (Fig.~\ref{fig:dis}).

\noindent\textbf{Two-round queries.} 
In order to give meaningful gradient updates to more queries during training, we propose to use two-round matching. After getting the first round matching $\sigma_1$ according to Eq.~\eqref{eq:smatch} through Hungarian algorithm~\cite{kuhn1955hungarian}, we do a second matching among the remaining queries to get $\sigma_2$ where we only use $\mathcal{L}_{ce}$ as the matching score. We simply assign the unmatched queries after two rounds to predict $\varnothing$. Thus, we easily include 2× matched queries as many as before into training by supervising them with informative ground truths. This design is time-efficient since we can get the second matching results directly by reusing the cost matrix $\mathcal{L}_{ce}$ from the first matching round.

\noindent\textbf{Hierarchical loss functions.} For $\sigma_1$ and $\sigma_2$, if we use the same mask loss $\mathcal{L}_{mask}$ on them, it will make the learned semantic features of $q_{\sigma_1(i)}$ and $q_{\sigma_2(i)}$ overlap. Then the total semantic variety of queries still remains limited.
So, we further design the hierarchical loss with different emphasizes on the two matched groups. For $\sigma_1$ which contains the best-fit queries towards ground truths, we expect queries to take primary responsibility to segment the whole objects. So, we use the \textbf{hard} mask classification loss $\mathcal{L}_{hard}$ for queries in $\sigma_1$:
\begin{equation}
    \label{eq:hard}
    \mathcal{L}_{hard} = \sum_{i=1}^{N^{gt}} [-\log \hat{p}_{\sigma_1(i)}(c_i) + \mathcal{L}_{mask}^{\sigma_1}(\hat{m}_{\sigma_1(i)}, m_i)],
\end{equation}
where $\mathcal{L}_{mask}^{\sigma_1}$ includes the weighting hyper-parameters $\lambda_{ce}^{\sigma_1}$ and $\lambda_{dice}^{\sigma_1}$ as in Eq.~\eqref{eq:smatch}.

For $\sigma_2$, each query as the second best-fit can also represent the matched object partially. Not required to learn all the semantic features of the object, instead we design the \textbf{soft} mask classification loss $\mathcal{L}_{soft}$ to help it refine its currently learned knowledge about that object.
$\mathcal{L}_{soft}$ has the similar formula as $\mathcal{L}_{hard}$ except using soft mask $sm_i$ to replace $m_i$ in Eq.~\eqref{eq:hard}:
\begin{equation}
    \begin{gathered}
    \mathcal{L}_{soft} = \sum_{i=1}^{N^{gt}} [-\log \hat{p}_{\sigma_2(i)}(c_i) + \mathcal{L}_{mask}^{\sigma_2}(\hat{m}_{\sigma_2(i)}, sm_i)], \text{where}~sm_i = \hat{m}_{\sigma_2(i)} \circ m_i, 
    \end{gathered}
\end{equation}
where $\circ$ is the Hadamard product. The soft mask is generated from the intersection area between the matched ground truth mask and the predicted mask, which pushes the query to focus on consolidating its learned semantic features on the matched object.

In our final hierarchical loss $\mathcal{L}_{hie}$ used by~\ourmodel, we balance the two mask terms in $\mathcal{L}_{hard}$ and $\mathcal{L}_{soft}$ by a round weight $\alpha$ and merge all the classification terms from three query groups, \ie, hard group, soft group, and unmatched group, into one classification term $\mathcal{L}_{cls}$:
\begin{equation}
\begin{split}
    \mathcal{L}_{hie} & = \sum_{i=1}^{N^{gt}} [\mathcal{L}_{mask}^{\sigma_1}(\hat{m}_{\sigma_1(i)}, m_i)    + \alpha\mathcal{L}_{mask}^{\sigma_2}(\hat{m}_{\sigma_2(i)}, sm_i)] + \mathcal{L}_{cls}.
\end{split}
\end{equation}
Our hierarchical loss introduces the hierarchical structures in queries by considering different semantic requirements in terms of the fitness of object queries to ground truths and optimizing the matching relationship between queries and objects, which is critical for richer expression ability and higher accuracy.

\subsection{Temporal Aggregation Decoder}
\label{section:deco}
\begin{wrapfigure}[15]{l}{.58\linewidth}
    \centering
    \includegraphics[width=\linewidth]{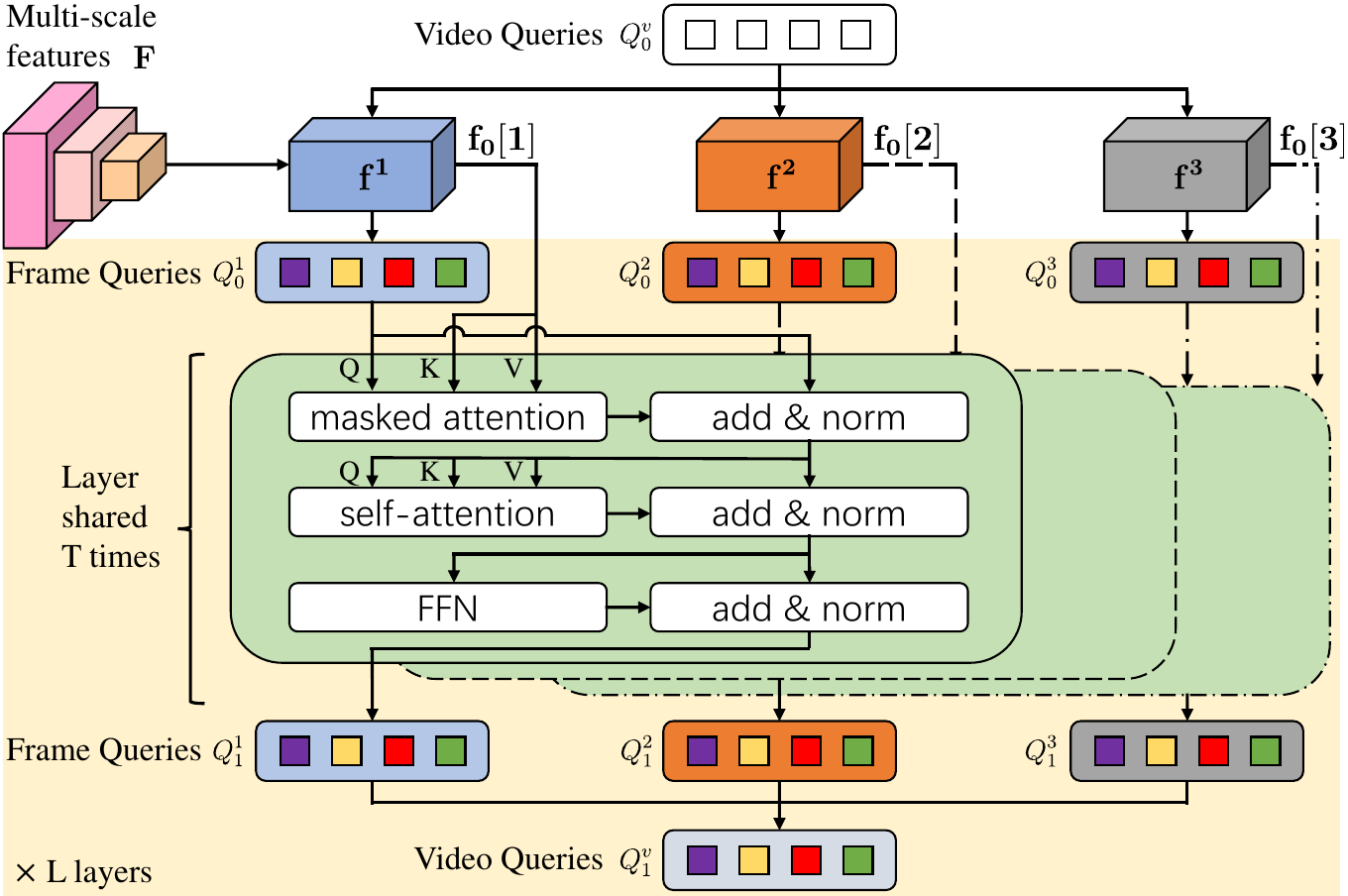}
    \caption{The architecture of TAD.}
    \label{fig:tad}
\end{wrapfigure}
Since VSS requires detecting objects across frames, it is vital to learn the semantic information from multiple frames jointly for temporal consistency.
Denote the multi-scale features from pixel decoder as $\mathbf{F} = \{\mathbf{f_i} | \mathbf{f_i} \in \mathbb{R}^{T \times H_i \times W_i \times C}\}_{i=1}^n$, where $\mathbf{f_i[t]} \in \mathbb{R}^{H_i \times W_i \times C}$ indicates the $i$-th scale features of $t$-frame. One naive way to model the temporal context is to directly attend queries to the features of $T$ frames $\mathbf{f_i}$. But the $T$ frames token sequence is very long and thus it is very hard to learn the temporal information. 

To alleviate the issue, we propose the lightweight \textbf{Temporal Aggregation Decoder (TAD)} as in Fig.~\ref{fig:tad} to effectively model the temporal interactions while doing the attention at the frame level. 
In this decoder, we initiate one set of video-level queries $Q^v_0=\{q^v_{0,i}\}_{i=1}^{N} \in \mathbb{R}^{N \times C}$ which is responsible for video-level features. We flatten and concatenate the multi-scale features $\{\mathbf{f_i}\}_{i=1}^n$ to form frame features $\{\mathbf{f^t} | \mathbf{f^t} \in \mathbb{R}^{(\sum_{i=1}^n H_iW_i) \times C}\}_{t=1}^T$. Then before passing into the decoder layer, $Q^v_0$ will attend to the frame features $\mathbf{f^t}$ to produce the frame-level queries $Q^t_0=\{q^{t}_{0,i}\}_{i=1}^{N} \in \mathbb{R}^{N \times C}$ which are only responsible to interact with the $t$-th frame:
\begin{equation}
\label{eq:subq}
    Q^{t}_0 = \text{SM}(\frac{f_Q(Q^v_0) f_K(\mathbf{f^t})^T}{\sqrt{C}})f_V(\mathbf{f^t}),
\end{equation}
where $\text{SM}$ means $\text{Softmax}$, $f_Q, f_K,$ and $f_V$ are linear maps to generate query, key, and value respectively. For simplicity, we omit the formulation of multi-head attention~\cite{NIPS2017_3f5ee243} here.

Then we pass one resolution of the multi-scale features $\mathbf{f_l}$ into $l$-th decoder layer at a time in a round-robin fashion~\cite{cheng2022masked}. In each layer, we do the masked attention at the frame level:
\begin{equation}
    \begin{gathered}
        Q^{t}_l = Q^{t}_{l-1} + \text{SM}(\mathcal{M}_{l-1}^t + \frac{f_Q(Q^{t}_{l-1}) f_K(\mathbf{f_l[t]})^T}{\sqrt{C}})f_V(\mathbf{f_l[t]}),  \quad \mathcal{M}_{l-1}^t = 
        \begin{cases}
          0 & \text{if $M_{l-1}^t(x,y)$ = 1} \\
          -\infty & \text{otherwise}
        \end{cases}
    \end{gathered}
\end{equation}
where $Q^{t}_l$ denotes the frame-level queries at $l$-th layer, and $M_{l-1}^t \in \{0,1\}^{N \times H_lW_l}$ is the mask prediction from $Q^{t}_{l-1}$ resized to the resolution of $\mathbf{f_l[t]}$. 
After this frame-level attention of each layer, we aggregate all frame-level queries $q^{t}_{l,i}$ to our video-level queries $q^{v}_{l,i}$:
\begin{equation}
    q^{v}_{l,i} = q^{v}_{l-1,i} + \frac{\sum_{t=1}^T q^{t}_{l,i} \times \text{exp}(\text{FC}(q^{t}_{l,i}))}{\sum_{t=1}^T \text{exp}(\text{FC}(q^{t}_{l,i}))},
\end{equation}
where FC is a fully connected layer to reduce channels to $1$ and a softmax is used to obtain the weight on each $q^{t}_{l,i}$ over $T$ frames. 
The aggregation operation enables video-level queries to collect frame-level information and learn the globally temporal-sensitive representations. 

During training, we combine the class predictions from video-level queries and mask predictions from frame-level queries as input $\{\hat{y}_i=(\hat{p}_i^v,\{\hat{m}_i^t\}_{t=1}^{T})\}_{i=1}^{N}$ (ignore $l$ for simplicity) to the two-round matching module to get both groups involved in the matching phase. For inference, we simply use the average of predictions from video-level and frame-level queries:
\begin{equation}
    \hat{y}_{t,h,w} = \argmax_{c\in \mathbb{C}} \sum_{i=1}^N \frac{1}{2}(\hat{p}_i^v(c)\cdot \hat{m}^v_{i,t,h,w} + \hat{p}^{t}_i(c)\cdot \hat{m}^{t}_{i,h,w}).
\end{equation}
In this way, we could avoid the long attention sequence and model the temporal information through the interactions between video-level and frame-level queries effectively.
\section{Experiments}

\subsection{Implementation Details}
\ourmodel~is implemented using MMSegmentation toolbox~\cite{contributors2020mmsegmentation}. 
Following~\cite{sun2022coarse}, we use the encoder of SegFormer~\cite{xie2021segformer} as our backbones, which is a hierarchical transformer pretrained on ImageNet~\cite{krizhevsky2017imagenet}. The pixel decoder and transformer decoder follow Mask2Former~\cite{cheng2022masked}.

For our hierarchical loss, we set $\lambda_{ce}^{\sigma_1} = \lambda_{dice}^{\sigma_1} = 2.5, \lambda_{ce}^{\sigma_2} = \lambda_{dice}^{\sigma_2} = 0.25$ and $\alpha = 0.5$. We use AdamW~\cite{loshchilov2017decoupled} and the \textit{poly}~\cite{chen2017deeplab} learning rate schedule with an initial learning rate of $10^{-4}$ and a weight decay of $0.05$. A learning rate multiplier of $0.1$ is applied to our backbone. 
We adopt data augmentations including standard random scale jittering with a resizing scale sampled from 0.5 to 2.0 followed by random cropping to 480$\times$480, random horizontal flipping with probability 0.5, and standard random color jittering. We calculate $\mathcal{L}_{mask}$ with sampled points in both matching scores and final losses following~\cite{cheng2022masked}. 
We train models using 4 NVIDIA GPUs with a batch size of 8 (2 clips per GPU). 
For 1-frame clips training ($t=1$), we use the first 30k as warm-up iterations with the original one-round matching and loss. After 30k, we apply our hierarchical loss for 130k iterations. For multiple-frame clips training ($t>1$), we replace the original transformer decoder with TAD, load the weights from the $t=1$ model, and finetune TAD by freezing the backbone and pixel decoder. The finetune setting is the same as above except using only 12K iterations which are very fast.
For testing, we resize all frames on VSPW to 480$\times$853 and conduct single-scale inference. \ourmodel~is  flexible to infer a video of arbitrary length without any post-processing. By default, we divide one video into non-overlapping clips as long as the training clips. For results, we report the mean of three runs.

\begin{table*}[!t]
\centering
\small
\begin{tabularx}{\linewidth}{c|c|c|c|c}
\hline
Method  & Backbone  & Params (M) $\downarrow$ & mIoU $\uparrow$ & Weighted IoU $\uparrow$ \\ \hline
DeepLabv3+~\cite{chen2017rethinking} & ResNet-101 & 62.7 &34.7  & 58.8 \\ 
UperNet~\cite{xiao2018unified} & ResNet-101 & 83.2 &36.5  & 58.6 \\ 
PSPNet~\cite{zhao2017pyramid} & ResNet-101 & 70.5 &36.5  & 58.1 \\ 
OCRNet~\cite{yuan2020object} & ResNet-101 & 58.1 &36.7  & 59.2 \\ 
ETC~\cite{liu2020efficient} & PSPNet & 89.4 &36.6  & 58.3 \\ 
ETC~\cite{liu2020efficient} & OCRNet & 58.1 &37.5  & 59.1 \\ 
NetWarp~\cite{xiao2018unified} & PSPNet & 89.4 &37.0  & 57.9 \\ 
NetWarp~\cite{xiao2018unified} & OCRNet & 58.1 &37.5  & 58.9 \\ 
TCB\textsubscript{st-ppm} ~\cite{miao2021vspw} & ResNet-101 & 70.5 &37.5  & 58.6 \\ 
TCB\textsubscript{st-ocr} ~\cite{miao2021vspw} & ResNet-101 & 58.1 &37.4 &  59.3\\ 
TCB\textsubscript{st-ocr-mem} ~\cite{miao2021vspw} & ResNet-101 & 58.1 &37.8 &  59.5\\ 
Video K-Net (Deeplabv3+)~\cite{li2022video} & ResNet-101 & \_ & 37.9 & \_ \\
Video K-Net (PSPNet)~\cite{li2022video} & ResNet-101 & \_ & 38.0 & \_\\
SegFormer~\cite{xie2021segformer} & MiT-B1 & 13.8 & 36.5 & 58.8  \\ 
SegFormer~\cite{xie2021segformer} & MiT-B2 & 24.8 &43.9  & 63.7 \\ 
CFFM ($t=4$)~\cite{sun2022coarse} & MiT-B1 & 15.5 & 38.5 & 60.0 \\ 
CFFM ($t=4$)~\cite{sun2022coarse} & MiT-B2 & 26.5 &44.9  & 64.9 \\ 
MRCFA ($t=4$)~\cite{sun2022mining} & MiT-B1 & 16.2 & 38.9 & 60.0 \\ 
MRCFA ($t=4$)~\cite{sun2022mining} & MiT-B2 & 27.3 & 45.3 & 64.7 \\ 
\hline
\multirow{3}{*}{Mask2Fomer ($t=1$)~\cite{cheng2022masked}} & MiT-B0 &  23.0 & 38.9  &  60.9 \\ 
 & MiT-B1 &  33.0 & 43.3 &  63.6 \\ 
 & MiT-B2 & 44.0  & 47.6  &  65.4 \\ 
\hline
\multirow{3}{*}{\ourmodel~($t=1$)} & MiT-B0 &  23.0 & \textbf{39.8} & \textbf{61.3} \\ 
 & MiT-B1 & 33.0  & \textbf{44.1}  &  \textbf{64.2} \\ 
 & MiT-B2 &  44.0 & \textbf{48.5} & \textbf{66.2} \\ 
\hline 
\ourmodel~($t=2$) & MiT-B5 &  104.5 & \textbf{52.1} & \textbf{67.2} \\ 
\hline
\end{tabularx}
\caption{\textbf{Comparison with state-of-the-art methods on the VSPW validation set.} Our model outperforms both the best per-pixel classification approaches and the strong mask classification-based baseline.}
\label{table:results_baselines}
\end{table*}

\begin{table*}[!t]
    \begin{minipage}[c]{.5\linewidth}
        \centering
        \small
        \begin{tabular}{cccc}
        \toprule
         backbone & $t = 1$  & $t = 2$  & $t = 4$ \\ \hline
          MiT-B0 &    39.76 & 39.94 & \textbf{40.68} \\
          MiT-B1 &    44.06  & 44.68 & \textbf{45.19} \\
          MiT-B2 &  48.53  &  48.99 & \textbf{49.11} \\
        \bottomrule
        \end{tabular}
        \caption{Effects of training clip length.}
            \vspace{-0.4cm}
        \label{table:clip}
    \end{minipage}\hfill
    \begin{minipage}[c]{.5\linewidth}
        \centering
        \small
        \begin{tabular}{ccc}
        \toprule
         temporal setting & $t = 2$  & $t = 4$ \\ \hline
          one-to-video & 43.40   & 43.90  \\
          one-to-frame &  44.41  & 44.84  \\
          video-frame &  \textbf{44.68}   & \textbf{45.19}  \\
        \bottomrule
        \end{tabular}
        \caption{Temporal aggregation ablation.}
                    \vspace{-0.4cm}
        \label{table:tem}
    \end{minipage}
\end{table*}

\subsection{Main Results}
The comparison of \ourmodel~with state-of-the-art methods on VSPW~\cite{miao2021vspw} are listed in Table~\ref{table:results_baselines}. From DeepLabv3+~\cite{chen2017rethinking} to MRCFA~\cite{sun2022mining}, they are all per-pixel classification paradigms. We select the mask-classification-based Mask2Fomer~\cite{cheng2022masked} as the baseline model. From Table~\ref{table:results_baselines}, we can observe that the mask-classification-based models perform much better than previous per-pixel classification paradigms. The baseline model Mask2Fomer trained on 1-frame clips outperforms MRCFA trained on 4-frame clips with backbone MiT-B1 and MiT-B2 by 4.4\% and 2.3\% mIoU, respectively. 
It shows the huge potential of the mask-classification-based model over the per-pixel classification paradigm for the VSS task.

In the comparison between \ourmodel~($t=1$) and the baseline, \ourmodel~($t=1$) consistently boosts Mask2Fomer by a significant margin on both mIoU (0.8-0.9\%) and wIoU (0.4-0.8\%) without any extra-processing steps while having the same amount of parameters. Note that \ourmodel~($t=1$) does not use the TAD since the one-frame training clips do not provide temporal information. So, the only difference between \ourmodel~($t=1$) and Mask2Fomer lies in the hierarchical query design. 
It shows that our parameter-free hierarchical design can better utilize object queries for learning richer semantic features and achieve superior performance  effectively. 
We further use a larger backbone (MiT-B5) in \ourmodel~($t=2$) with TAD trained on two-frame clips to learn temporal clues. It sets the new state-of-the-art results of 52.1 mIoU and 67.2 wIoU.

\begin{wraptable}[6]{r}{.55\linewidth}
    \centering
    \small
    \begin{tabular}{cccc}
    \hline
    Matching & Loss  &  mIoU $\uparrow$ & wIoU $\uparrow$ \\ \hline
    one round & original loss & 43.26 & 63.56 \\
    two round & original loss & 43.70 & 63.77 \\
    two round & hierarchical loss & \textbf{44.06} & \textbf{64.16} \\ 
    \hline
    \end{tabular}
    \caption{Ablation study on hierarchical loss.}
    \label{table:ab_loss}
\end{wraptable}

\subsection{Ablation Studies}
Here, we provide ablation studies for more insights into the effects of different designs. The experiments are performed on VSPW~\cite{miao2021vspw} validation set using a MiT-B1 backbone.

\noindent\textbf{Training clip length.} Table~\ref{table:clip} (mIoU reported) shows how the length of training clips affects the performance of \ourmodel. When training on multiple-frame clips, we use the TAD to mine temporal information. On all three backbones, TAD clearly brings us  better results when using longer training clips that contain richer semantic context to help segmentation. It demonstrates the effectiveness of the proposed TAD to learn temporal clues from long clips.

\noindent\textbf{Temporal aggregation.} In Table~\ref{table:tem} (mIoU reported), we present ablation on different temporal learning settings of \ourmodel. Here, one-to-video means one set of object queries attends to the features of all frames. One-to-frame represents attending one set of queries to each frame by replicating that query set. And video-frame means TAD where we use the interaction between two sets of queries, \ie, video-level and frame-level queries, to aggregate temporal information. One-to-frame yields 1.11\% ($t=2$) and 0.94\% ($t=4$) mIoU improvements than one-to-video. It shows that directly attending queries to all frames is difficult to learn temporal clues due to the long token sequence of multiple frames. Change to video-frame setting with only 3.2M extra parameters involved further improves mIoU by 0.27\% ($t=2$) and 0.35\% ($t=4$), suggesting the separation of attention and explicitly modeling of frame aggregation are key factors for temporal learning.

\noindent\textbf{Hierarchical queries.} In Table~\ref{table:ab_loss}, we ablate our hierarchical query design to verify its effectiveness. The models are trained by 1-frame clips. We develop a variant by adding second-round matching and using the same original (hard) loss to the second-matched group. As mentioned earlier, simply adding one more matching round could involve more queries in the training, but the same loss applied to the two matched groups still limits the full utilization of all queries.
When we apply our proposed hierarchical loss, it brings a large improvement of 0.8\% mIoU and 0.6\% wIoU to the baseline model. It shows that our proposed parameter-free hierarchical design can effectively supervise the two query groups with different focuses, and the hierarchy learned inside the object queries is helpful for performance gains.

\section{Conclusion}
In this paper, we rethink the application of mask-classification-based models in the VSS domain. Based on the fact of low utilization of object queries during training, we present \ourmodel, a simple and strong mask-classification-based model for VSS. \ourmodel~is the first model to renovate the traditional one-to-one matching and introduce hierarchical structures into queries to fully exploit the representation ability of queries. 
Besides, we propose TAD to explicitly model the temporal interactions for cross-frame learning.
Equipped with the parameter-free hierarchical design optimizing the matching relationship between queries and objects, and TAD optimizing the interactions between queries and multiple frames, \ourmodel~achieves the state-of-the-art results on the VSPW dataset. 
We hope our methods can inspire future research in the segmentation domain. 

\bibliography{egbib}
\end{document}